\documentclass{article}
\usepackage{spconf,amsmath,graphicx,hyperref}
\usepackage{amsfonts}
\usepackage{booktabs}
\usepackage{multirow} 
\usepackage{makecell}
\usepackage{xcolor}
\usepackage{dcolumn}
\usepackage{hyperref}

\title{PSGait: Gait Recognition using Parsing Skeleton}
\name{\fontsize{11}{12}\selectfont Hangrui Xu$^{1, 2}$, Zhengxian Wu$^{1}$, Chuanrui Zhang$^{1}$, Zhuohong Chen$^{1}$, Zhifang Liu$^{1}$, Peng Jiao$^{1}$, Haoqian Wang$^{1}$$^{,}$\sthanks{Corresponding author: wanghaoqian@tsinghua.edu.cn}}
  
\address{$^{1}$The Shenzhen International Graduate School, Tsinghua University, China  \\   $^{2}$School of Computer Science and Information Engineering, Hefei University of Technology, China}

\begin{document}

\maketitle

\begin{abstract}
Gait recognition has emerged as a robust biometric modality due to its non-intrusive nature. Conventional gait recognition methods mainly rely on silhouettes or skeletons. While effective in controlled laboratory settings, their limited information entropy restricts generalization to real-world scenarios. To overcome this, we propose a novel representation called \textbf{Parsing Skeleton}, which uses a skeleton-guided human parsing method to capture fine-grained body dynamics with much higher information entropy. To effectively explore the capability of the Parsing Skeleton, we also introduce \textbf{PSGait}, a framework that fuses Parsing Skeleton with silhouettes to enhance individual differentiation. Comprehensive benchmarks demonstrate that PSGait outperforms state-of-the-art multimodal methods while significantly reducing computational resources. As a plug-and-play method, it achieves an improvement of up to 15.7\% in the accuracy of Rank-1 in various models. These results validate the Parsing Skeleton as a \textbf{lightweight}, \textbf{effective}, and highly \textbf{generalizable} representation for gait recognition in the wild. Code is available at \url{https://github.com/realHarryX/PSGait}.
\end{abstract}
\section{Introduction}
\label{sec:intro}

\begin{figure}[t]
\centering
\includegraphics[width=\linewidth]{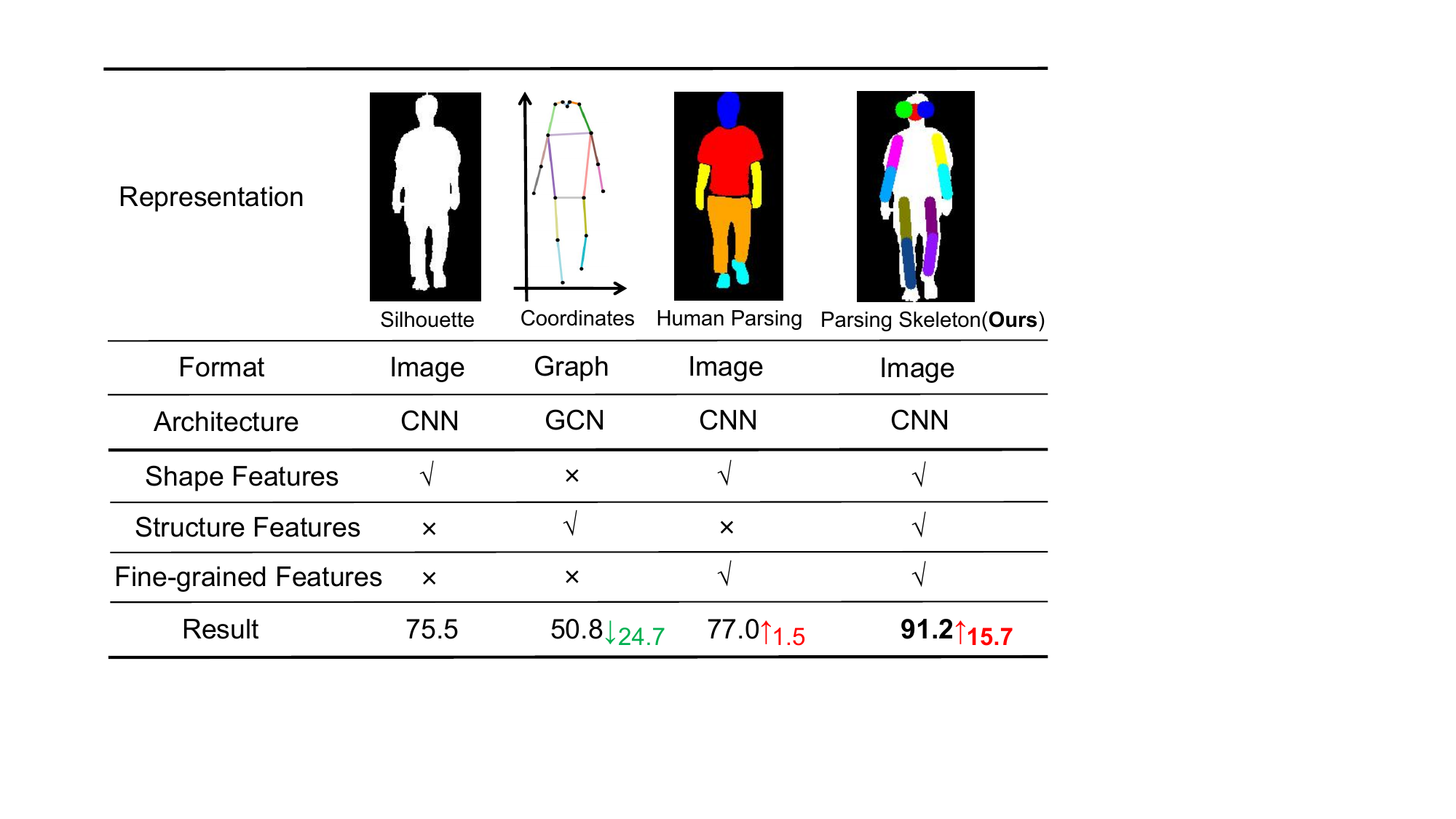}
\vspace{-6mm}
\caption{The comparison of various gait representations. Only a single frame is displayed. Result shows the Rank-1 on CCPG \cite{ccpg} for different representation inputs to GaitBase \cite{opengait}.}
\vspace{-4mm}
\label{fig:intro}
\end{figure}

Gait recognition identifies individuals from walking patterns. Compared with face or fingerprint, it is non-intrusive, contactless, and works at a distance without subject cooperation, making it attractive for identification, access control, and public-security use in the wild \cite{intro2, wu2016comprehensive, lmgait}.

Most methods rely on silhouettes \cite{yu2006framework}, GEIs \cite{han2005individual}, 2D \cite{gaitgraph} or 3D skeletons \cite{posegait}. Silhouettes are popular but suffer from low information entropy. They retain only coarse contours while discarding fine motion and structural details, and they are highly sensitive to clothing changes, occlusion, background noise, and background-subtraction or domain shifts. Moreover, they lack explicit body-part semantics \cite{zheng2023parsing}. Skeletons abstract motion via two-dimensional joint coordinates and have been modeled with GCNs \cite{gaitgraph, gaitgraph2, gaittr, gpgait}, yet still trail image-based methods. We attribute this gap to the sparsity and low dimensionality of joint sets, which limit spatial detail and long-range dependency modeling. Also, typical GCNs further suffer from restricted receptive fields.

\begin{figure*}[t]
\centering
\includegraphics[width=\linewidth]{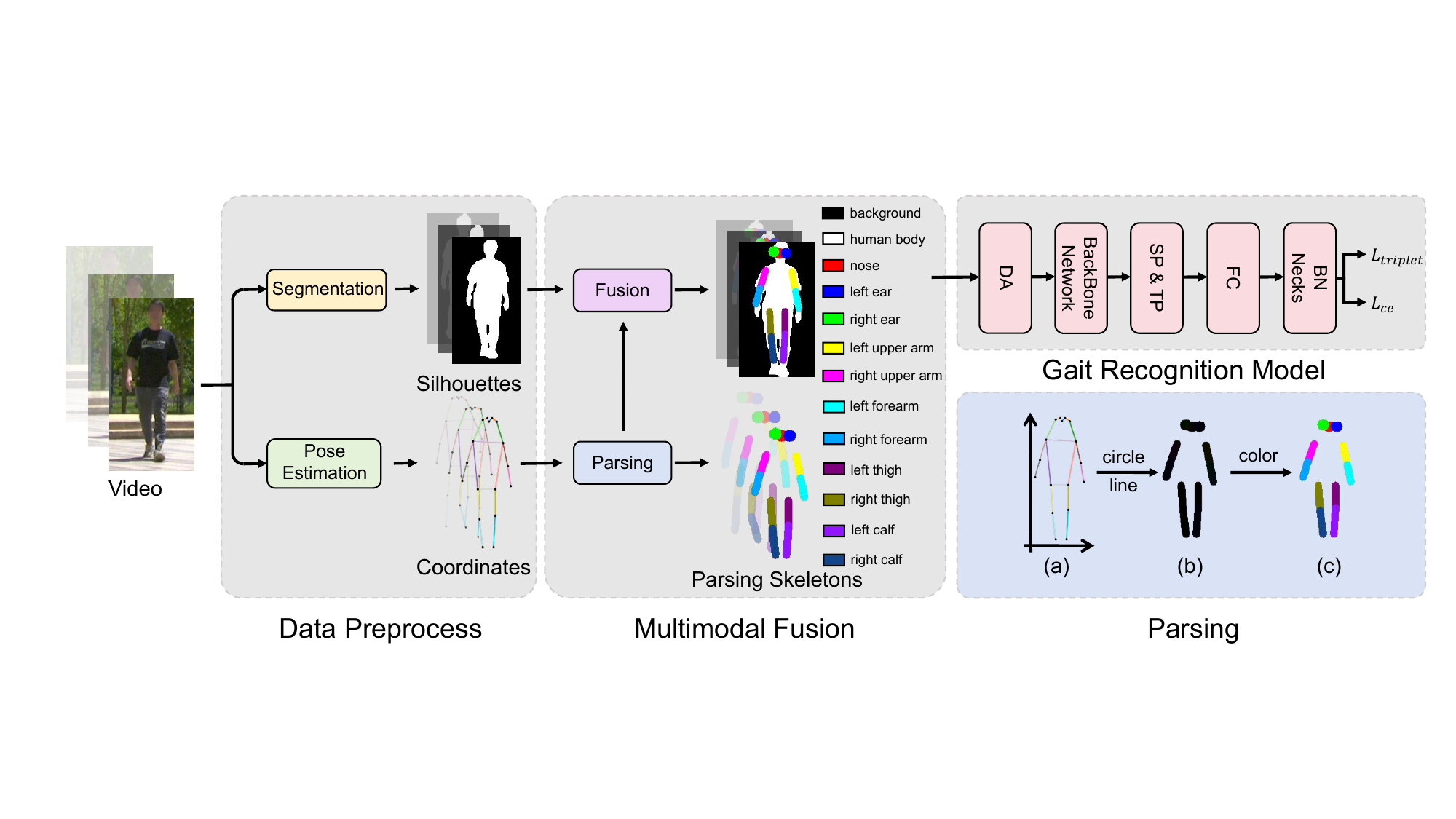}
\vspace{-8mm}
\caption{Overview of the PSGait framework. During data preprocessing, the silhouettes and skeletons are generated from video frames. After the parsing section, the Parsing Skeletons are fused with silhouettes. The fused data is then fed into the gait recognition model for individual differentiation. Both the data preprocessing and gait recognition models are interchangeable.}
\vspace{-4mm}
\label{fig:method}
\end{figure*}

To overcome single-modality limits, recent works fuse silhouettes and skeletons \cite{skeletongait, bifusion, cui2023multi, li2023transgait}. SkeletonGait++ \cite{skeletongait} visualizes joints as Gaussian points and attends with silhouettes, improving capacity but still missing inter-joint dynamics and part-level motion patterns—both crucial for discriminative gait.
Human parsing offers fine-grained part semantics and richer motion cues \cite{zheng2023parsing, wang2023gaitparsing, zheng2024takes, wang2023landmarkgait}, yet most parsing pipelines rely on RGB, which is sensitive to illumination, clothing, and occlusion, weakening robustness in the wild.

To address these challenges, we introduce \textbf{Parsing Skeleton}, a skeleton-guided human parsing representation that produces dense, part-aware body maps. It increases pixel-level information entropy with detailed motion cues, encodes explicit body-part semantics to separate diverse regional dynamics, and inherits skeletons’ robustness to environmental noise and domain shifts, thereby avoiding RGB fragility. Unlike sparse coordinates, Parsing Skeleton renders limbs and head as colored line segments and circles, boosting spatial density and preserving both local motion and global context. As an image representation, it pairs naturally with CNNs and fuses easily with silhouettes, while sidestepping the long-range and expressiveness issues of GCNs.

Building on this, we propose \textbf{PSGait}, which integrates Parsing Skeletons with silhouettes and feeds the fused sequences into gait models. Across multiple datasets, PSGait improves Rank-1 accuracy by up to 7.5\% over state-of-the-art multimodal methods with notably lower computation. Replacing silhouettes with our fused inputs yields up to 15.7\% Rank-1 gains across diverse backbones, highlighting Parsing Skeleton’s effectiveness for real-world gait recognition.

In summary, the contributions of this paper are:
(1) We propose Parsing Skeleton, a novel gait representation that offers higher information entropy and fine-grained body-part awareness.
(2) We introduce PSGait, a novel gait recognition framework that extracts comprehensive gait features with Parsing Skeletons and silhouettes to enhance the feature robustness in the wild.
(3) Extensive experiments on multiple datasets validate the \textbf{lightweight}, \textbf{effective} and \textbf{generalizable} nature of Parsing Skeleton. PSGait achieves state-of-the-art performance among multimodal gait recognition methods while maintaining high computational efficiency.

\section{Method}
\label{sec:method}

The proposed \textbf{PSGait} is illustrated in Fig.~\ref{fig:method}. First, during data preprocessing, the silhouette and skeleton for each frame of the gait video sequence are obtained through segmentation and pose estimation. Next, after the parsing section, the \textbf{Parsing Skeletons} are fused with silhouettes to obtain the multimodal gait representation. Finally, this multimodal gait representation with comprehensive information is fed into the gait recognition model to enhance its classification performance.

\subsection{Parsing Skeleton Representation}\label{AA}
In this section, we introduce the definition of the novel representation, i.e., Parsing Skeleton, for gait recognition.
Given a human walking sequence of frames $S=\{\mathbf{C}^i\}^N_{i=1}$, where $\mathbf{C}^i \in \mathbb{R}^{K \times 2}$ is one coordinates frame and $N$ is the length of the sequence, we can feed each frame $\mathbf{C}^i$ into the parsing section $P(\cdot)$ to obtain the Parsing Skeleton frame $\textbf{x}^i=P(\textbf{C}^i)$.
The parsing section $P(\cdot)$ assigns each pixel to one of the $K$ body-part classes, including the background, and renders it with the corresponding color $\textbf{C}^i$.
Therefore, the Parsing Skeleton sequence can be formulated as:
\begin{equation*}
\setlength{\abovedisplayskip}{3pt}
\setlength{\belowdisplayskip}{3pt}
X= \{ \mathbf{x}^i\}^{N}_{i=1},
\end{equation*}
where $\textbf{x}^{i}\in \mathbb{R}^{H \times W}$ is the Parsing Skeleton frame.
For each pixel $p^j$ of $\mathbf{x}^i$, we have $p^j \in \{0, 1, ..., K-1\}$.

Based on the above definition, we compare the proposed Parsing Skeleton representation with the binary silhouette from the viewpoint of information theory.
Inspired by the definition of information entropy proposed by Claude E. Shannon \cite{shannon1948mathematical}, the pixel-level information entropy of the Parsing Skeleton can be denoted as
\begin{equation*}
\setlength{\abovedisplayskip}{3pt}
\setlength{\belowdisplayskip}{3pt}
\mathcal{H} = - \sum_{k=0}^{K} p_k \cdot \log(p_k),
\end{equation*}
where $p_k$ is the possibility of one pixel belonging to the k-th class. For the conventional binary silhouette, each pixel only belongs to the foreground or the background, i.e., the $K=2$, so the information of the binary pixel is only one bit. The combination of Parsing Skeleton with silhouette, where $K=13$, leads to a significantly higher information entropy.
Therefore, we consider Parsing Skeleton as a gait representation with higher information entropy to encode the fine-grained dynamics of body parts during walking.

\begin{table*}[h]
\centering
\caption{Quantitative comparison with state-of-the-art gait recognition methods across three datasets. The best performances are indicated in \textbf{bold}, and the second-best methods are \underline{underlined}.}
\vspace{1mm}
\resizebox{2.0\columnwidth}{!}{
\setlength{\tabcolsep}{0.5em}
\begin{tabular}{c|c|c|ccccccccccc}
\toprule
\multirow{3}{*}{Input}      & \multirow{3}{*}{Method} & \multirow{3}{*}{Source} & \multicolumn{11}{c}{Testing Datasets}                                                                                     \\ \cline{4-14} 
                            &                         &                         & \multicolumn{5}{c|}{CCPG}                                & \multicolumn{4}{c}{Gait3D}  & \multicolumn{2}{c}{SUSTech1K}   \\ \cline{4-14} 
                            &                         &                         & CL & UP & DN & BG & \multicolumn{1}{c|}{Mean} & Rank-1 & Rank-5  & mAP  & \multicolumn{1}{c|}{mINP} & Rank-1 & Rank-5 \\ \midrule 
\multirow{2}{*}{\begin{tabular}[c]{@{}c@{}} Skeleton   \end{tabular}}   & GPGait \cite{gpgait}             & ICCV2023               & 54.8    & 65.6   & 71.1 & 65.4 & \multicolumn{1}{c|}{64.2}                          & 22.4   & \multicolumn{3}{c|}{-} & 41.5 & 65.4\\
                            & SkeletonGait \cite{skeletongait}                  & AAAI2024               & 40.4    & 48.5   & 53.0 & 61.7 & \multicolumn{1}{c|}{50.9}                          & 38.1    & 56.7 & 28.9 &\multicolumn{1}{c|}{16.1} & 50.1 & 72.6\\ \midrule

\multirow{4}{*}{Silhouette} & GaitSet \cite{gaitset}                 & TPAMI2022                & 60.2    & 65.2   & 65.1   & 68.5    & \multicolumn{1}{c|}{64.8}        & 36.7   & 58.3    & 30.0 & \multicolumn{1}{c|}{17.3} & 65.0 & 84.8 \\
                            & GaitPart \cite{gaitpart}                & CVPR2020                & 64.3    & 67.8   & 68.6   & 71.7    & \multicolumn{1}{c|}{68.1}       & 28.2   & 47.6    & 21.6 & \multicolumn{1}{c|}{12.4} & 59.2 & 80.8 \\
                            & GaitBase \cite{opengait}                & TPAMI2025                & 71.6    & 75.0   & 76.8                          & 78.6   & \multicolumn{1}{c|}{75.5} & 64.6   & \multicolumn{3}{c|}{-} & 76.1  & 89.4\\
                            & DeepGaitV2 \cite{deepgait}          & Arxiv2023               & 78.6    & 84.8   & 80.7   & 89.2    & \multicolumn{1}{c|}{83.3}       & 74.4   & 88.0    & 65.8 & \multicolumn{1}{c|}{-}  & 77.4 & 90.2  \\ 
                            \midrule 
\multirow{1}{*}{\begin{tabular}[c]{@{}c@{}}Parsing \end{tabular}} & GaitBase$^{p}$ \cite{opengait} & TPAMI2025 & 59.1 & 62.1 & 66.8 & 68.1 & \multicolumn{1}{c|}{64.0} & \multicolumn{4}{c|}{-} &71.7 &88.4
\\ \midrule
\multirow{4}{*}{\begin{tabular}[c]{@{}c@{}}Silhouette+\\ Skeleton \end{tabular}} 
& BiFusion \cite{bifusion}          & MTA2023               & 62.6    & 67.6   & 66.3   & 66.0    & \multicolumn{1}{c|}{65.6}       & \multicolumn{4}{c|}{-}  & 62.1 & 83.4 \\ 
& SkeletonGait++ \cite{skeletongait}         & AAAI2024               & 79.1    & 83.9   & 81.7   & 89.9    & \multicolumn{1}{c|}{83.7}       & 77.6   & 89.4    & 70.3 & \multicolumn{1}{c|}{42.6}  & 81.3 & \underline{95.5} \\
& PSGait-B      & Ours  & \textbf{87.1}   & \textbf{90.7}   & \textbf{92.8}   & \textbf{94.3}    & \multicolumn{1}{c|}{\textbf{91.2}}  & \underline{78.4}   & \underline{91.3}    & \underline{72.2} & \multicolumn{1}{c|}{\underline{49.3}}  & \underline{84.3} & 95.1  \\  
& PSGait-D      & Ours  & \underline{84.9}   & \underline{86.7}   & \underline{89.8}   & \underline{90.1}    & \multicolumn{1}{c|}{\underline{87.9}}  & \textbf{81.2}   & \textbf{91.6}    & \textbf{75.2} & \multicolumn{1}{c|}{\textbf{54.9}} & \textbf{86.1} & \textbf{95.9}    \\  \bottomrule
\end{tabular}
}
\vspace{-5mm}
\label{tab:results1}
\end{table*}

\subsection{Parsing Section}\label{AA}
To mitigate the impact of erroneous results from the pose estimation during preprocessing, only points with confidence \(c_i\) exceeding the threshold $\tau$ are used. Additionally, any points outside the valid region (\( H \times W \)) are disregarded. Therefore, a valid point (\(x_i, y_i\)) must satisfy the following condition:
\begin{equation*}
\setlength{\abovedisplayskip}{3pt}
\setlength{\belowdisplayskip}{3pt}
J' = \left\{ (x_i, y_i) \mid 0 \leq x_i < W, \ 0 \leq y_i < H, \ c_i \geq \tau \right\}.
\end{equation*}
According to the joint point configuration in COCO17, we selected the corresponding joint points for each body part. This results in the following mapping $M$ between human body joints \(k_i\) and their corresponding body parts \(P_j\):
\begin{equation*}
\setlength{\abovedisplayskip}{3pt}
\setlength{\belowdisplayskip}{3pt}
M(P_j) = \{k_1, k_2, \dots, k_m\}.
\end{equation*}

For the head representation, we introduce circles centered at the three key head points (\(x, y\)) with a specified radius \(r\). For the representation of each limb, a line of specified width is parsed on the image.
Thereby, as shown in Fig.~\ref{fig:method}, given sparse two-dimensional \textbf{(a)original coordinates}, we can get the dense \textbf{(b)skeleton image} to ensure that its information can be effectively captured by the CNN after reduced to a uniform size.
Additionally, different colors \(c\) are assigned to each body part to enhance the fine-grained information of the human body. Then, we can get the \textbf{(c)parsing skeleton}.

\subsection{Fusion Strategy}\label{BB}
We investigate two distinct fusion strategies for combining Parsing Skeleton and silhouette in representation learning. The first is \textbf{Composite Representation Fusion(CRF)}, which overlays the Parsing Skeleton onto the silhouette to yield a single image that jointly encodes global shape and local structure with minimal preprocessing. The second is \textbf{Disentangled Channel Fusion(DCF)}, which separates skeletal components into dedicated channels with the silhouette in an additional channel, offering a structured, component-aware representation. The former is simple and computationally light, promoting coherent spatial learning. The latter enhances part-level discriminability but increases input dimensionality and risks cross-channel spatial dilution. 

\subsection{Gait Recognition Model}\label{D}
Given a sequence of images $\mathbf{s}$, a backbone $\mathcal{B}$ extracts features
\begin{equation*}
\setlength{\abovedisplayskip}{3pt}
\setlength{\belowdisplayskip}{3pt}
\mathbf{f}=\mathcal{B}(\mathbf{s}) \in \mathbb{R}^{n\times c \times s \times h \times w},
\end{equation*}

where $n,c,s,h,w$ denote frame count, channels, sequence length, height, and width. Temporal Pooling (TP) \cite{gaitset} aggregates temporal cues to
\begin{equation*}
\setlength{\abovedisplayskip}{3pt}
\setlength{\belowdisplayskip}{3pt}
z=TP(\mathbf{f}) \in \mathbb{R}^{n \times c \times h \times w}.
\end{equation*}

Horizontal Pooling (HP) \cite{fu2019horizontal} partitions $z$ into horizontal stripes and applies global pooling per stripe:
\begin{equation*}
\setlength{\abovedisplayskip}{3pt}
\setlength{\belowdisplayskip}{3pt}
\mathbf{f}^{\prime}=\text{maxpool}(z_s)+\text{avgpool}(z_s).
\end{equation*}

A fully connected layer followed by BNNeck \cite{bnneck} maps features to a metric space, optimized by cross-entropy:
\begin{equation*}
\setlength{\abovedisplayskip}{3pt}
\setlength{\belowdisplayskip}{3pt}
L_{ce}=-\sum_{i=1}^{n} y_i \log(\widehat{y}_i).
\end{equation*}

To enhance inter/intra-class separation, we employ triplet loss \cite{schroff2015facenet} over $N$ triplets $(\mathbf{s}_i^a,\mathbf{s}_i^p,\mathbf{s}_i^n)$:
\begin{equation*}
\setlength{\abovedisplayskip}{3pt}
\setlength{\belowdisplayskip}{3pt}
L_{\text{triplet}}=\sum_{i=1}^{N}\max\Bigl(0,\ \|f(\mathbf{s}_i^a)-f(\mathbf{s}_i^p)\|_2^2-\|f(\mathbf{s}_i^a)-f(\mathbf{s}_i^n)\|_2^2+\alpha\Bigr),
\end{equation*}

where $f(\cdot)$ is the embedding function and $\alpha$ is the margin. The overall objective function is a combination of the cross-entropy loss and the triplet loss.
This design yields compact, stripe-aware representations and jointly optimizes identification and metric discrimination.

\section{EXPERIMENTS}
\label{sec:experiments}

\subsection{Datasets and Implementation Details}\label{AA}
We evaluated our proposed method on three commonly used datasets: SUSTech1K \cite{sustech1k}, CCPG \cite{ccpg} and Gait3D \cite{gait3d}. Our experiments strictly follow the official evaluation protocols.
Since skeleton data is unavailable for CCPG, we employ Sapiens \cite{sapiens} for 2D pose estimation, while original coordinates are used for SUSTech1K and Gait3D in COCO17 format.
To evaluate generalization, Parsing Skeletons are fed into two state-of-the-art gait recognition models: GaitBase \cite{opengait} and DeepGaitV2 \cite{deepgait}. For data augmentation (DA), we maintain the same setup as in OpenGait and apply the DAGait \cite{wu2025dagait} method. The fused inputs are resized to \( 64 \times 44 \) before being fed into the recognition models. All other settings are consistent with OpenGait \cite{opengait}.

\vspace{-4mm}
\begin{table}[h]
\centering
\caption{Generalization experiments on two baseline models across three datasets. PS: Parsing Skeleton Method}
\label{tab:gen1}
\setlength{\tabcolsep}{0.5em}
\resizebox{\columnwidth}{!}{%
\begin{tabular}{c|cc|cc|c}
\toprule
\multirow{2}{*}{Method} & \multicolumn{2}{c}{SUSTech1K} & \multicolumn{2}{|c|}{Gait3D} & CCPG \\
& Rank-1 & Rank-5 & Rank-1 & mAP & Rank-1 \\
\midrule
GaitBase \cite{opengait} & \makecell[l]{76.1} & \makecell[l]{89.4} & \makecell[l]{64.6} & \makecell[l]{54.5} & \makecell[l]{75.5} \\ GaitBase+PS & 84.3{\scriptsize \textcolor{red}{\textbf{+8.2}}} & 95.1{\scriptsize \textcolor{red}{\textbf{+5.7}}} & 78.4{\scriptsize \textcolor{red}{\textbf{+13.8}}} & 72.2{\scriptsize \textcolor{red}{\textbf{+17.7}}} & 91.2{\scriptsize \textcolor{red}{\textbf{+15.7}}} \\ \midrule DeepGaitV2 \cite{deepgait} & \makecell[l]{77.4} & \makecell[l]{90.2} & \makecell[l]{74.4} & \makecell[l]{65.8} & \makecell[l]{83.3} \\ DeepGaitV2+PS & 86.1{\scriptsize \textcolor{red}{\textbf{+8.7}}} & 95.9{\scriptsize \textcolor{red}{\textbf{+5.7}}} & 81.2{\scriptsize \textcolor{red}{\textbf{+6.8 }}} & 75.2{\scriptsize \textcolor{red}{\textbf{+9.4 }}} & 87.9{\scriptsize \textcolor{red}{\textbf{+4.6 }}} \\ \bottomrule
\end{tabular}%
}
\vspace{-6mm}
\end{table}

\subsection{Comparison with State-of-the-art Methods}

PSGait-B and PSGait-D use GaitBase and DeepGaitV2 as gait recognition models, respectively. As shown in Table~\ref{tab:results1}, PSGait is consistently superior to existing gait recognition approaches across multiple benchmarks and input modalities, demonstrating strong robustness under challenging conditions. These results confirm that PSGait effectively captures part-level dynamics and global structures, offering an accurate and practical solution for real-world gait recognition.

\vspace{-4mm}
\begin{table}[h]
\centering
\caption{Cross-domain experiments: models are trained on CCPG and tested on SUSTech1K and Gait3D datasets.}
\vspace{1mm}
\resizebox{0.9\columnwidth}{!}{
\setlength{\tabcolsep}{0.5em}
\begin{tabular}{c|cccc}
\toprule
\multirow{3}{*}{Method}     & \multicolumn{4}{c}{Testing Datasets}                                                                                     \\ \cline{2-5} 
                            & \multicolumn{2}{c|}{SUSTech1K}                                & \multicolumn{2}{c}{Gait3D}     \\ 
                            & Rank-1 & \multicolumn{1}{c|}{Rank-5} & Rank-1 & Rank-5 \\ \midrule 
GaitBase \cite{opengait}              & \makecell[l]{16.8}    & \multicolumn{1}{c|}{\makecell[l]{39.2}}   & \makecell[l]{11.8} & \makecell[l]{22.6}     \\
GaitBase+PS             & 45.9{\scriptsize \textcolor{red}{\textbf{+29.1}}}    & \multicolumn{1}{c|}{71.9{\scriptsize \textcolor{red}{\textbf{+32.7}}}}   
                        & 44.0{\scriptsize \textcolor{red}{\textbf{+32.2}}}    & 59.1{\scriptsize \textcolor{red}{\textbf{+36.5}}}     \\
                            \bottomrule

\end{tabular}
}
\label{tab:gen2}
\vspace{-6mm}
\end{table}

\subsection{Generalization of the Parsing Skeleton}

We further evaluate the generalization ability of the proposed Parsing Skeleton (PS) on different benchmarks and training settings. As shown in Table~\ref{tab:gen1} and ~\ref{tab:gen2}, incorporating PS consistently boosts the performance of both GaitBase and DeepGaitV2 across multiple datasets. In the within-domain setting, PS significantly improves recognition accuracy, demonstrating that it provides complementary structural cues to silhouettes. In the cross-domain setting, where models are trained on CCPG and tested on other datasets, PS brings substantial gains under domain shifts, highlighting its robustness to varying environments. These results confirm that Parsing Skeleton is a \textbf{plug-and-play} and \textbf{generalizable} representation. It consistently improves performance across architectures and datasets, making it well-suited for real-world deployment in gait recognition systems.

\begin{table}[h]
\centering
\caption{Efficiency and accuracy comparison between PSGait-B and the SOTA method using four RTX 2080 GPUs.}
\resizebox{\columnwidth}{!}{
\setlength{\tabcolsep}{0.5em}
\begin{tabular}{c|cc|cc|cc}
\toprule
\multirow{2}{*}{Method} & Params & \multicolumn{1}{c|}{GPU} & \multicolumn{1}{c}{Preproc.} & \multicolumn{1}{c|}{Train} & SUSTech1K & CCPG\\ 
& (M) & (GB) & (ms/f) & (s/ep) & Rank-1 & Rank-1 \\ \midrule
SkeletonGait++ & \makecell[l]{9.13} & \makecell[l]{5.14} & \makecell[l]{0.71} & \makecell[l]{0.34} & \makecell[l]{81.3} & \makecell[l]{83.7} \\
PSGait-B(Ours) & 8.02{\scriptsize \textcolor{red}{\textbf{-12\%}}} & 4.70{\scriptsize \textcolor{red}{\textbf{-9\%}}} & 0.47{\scriptsize \textcolor{red}{\textbf{-33\%}}} & 0.27{\scriptsize \textcolor{red}{\textbf{-21\%}}} & \makecell[l]{84.3{\scriptsize \textcolor{red}{\textbf{+3\%}}}} & 91.2{\scriptsize \textcolor{red}{\textbf{+8\%}}} \\
\bottomrule
\end{tabular}
}
\label{tab:eff}
\vspace{-6mm}
\end{table}

\subsection{Lightweight of the PSGait}
To assess efficiency, we compare PSGait-B with SkeletonGait++ in terms of model size, computational cost, and recognition accuracy. As shown in Table~\ref{tab:eff}, PSGait-B requires fewer parameters and less GPU memory, while also reducing preprocessing and training time. Despite its lower complexity, PSGait-B achieves higher accuracy on both SUSTech1K and CCPG, surpassing SkeletonGait++. These results demonstrate that PSGait is not only accurate but also \textbf{lightweight}, making it well-suited for real-world deployment.

\vspace{-5mm}
\begin{table}[h]
\centering
\caption{Radius/Width ablation using GaitBase, SUSTech1K.  }
\vspace{1mm}
\resizebox{0.6\columnwidth}{!}{
    \setlength{\tabcolsep}{0.5em}
        \begin{tabular}{cc|cc}
        \toprule
          Radius   & Width      & Rank-1 & Rank-5     \\ \midrule
         3 & 3              & 79.6  & 92.9  \\
         10 & 12          & \textbf{84.3}  & \textbf{95.1}  \\
         20 & 24                    & 82.7  & 94.8  \\
        \bottomrule
        \end{tabular}
        }
\label{tab:Ablation1}
\vspace{-6mm}
\end{table}

\vspace{-4mm}
\begin{table}[h]
\centering
\caption{Ablation on fusion strategy using GaitBase, CCPG.  }
\vspace{1mm}
\resizebox{0.75\columnwidth}{!}{
    \setlength{\tabcolsep}{0.5em}
        \begin{tabular}{c|ccccc}
        \toprule
          Strategy   &   CL    & UP & DN  & BG & Mean    \\ \midrule
            DCF  & 83.8              & 88.1  & 89.4  & 91.9 & 88.3\\ 
         CRF & \textbf{87.1}      & \textbf{90.7}  & \textbf{92.8} & \textbf{94.3} & \textbf{91.2}\\
        \bottomrule
        \end{tabular}
        }
\label{tab:Ablation2}
\vspace{-7mm}
\end{table}

\subsection{Ablation Study}
We conduct ablation experiments to examine the effect of different design choices in PSGait. Table~~\ref{tab:Ablation1} evaluates the influence of circle radius and line width in the Parsing Skeleton. The results show that moderate settings achieve the best balance. In contrast, too small values fail to capture sufficient structural information, while overly large radius or thickness cause occlusion of key details, leading to performance degradation. 
Table~~\ref{tab:Ablation2} compares two fusion strategies. The proposed CRF consistently outperforms DCF across all probe conditions, leading to higher mean accuracy. This demonstrates that directly integrating Parsing Skeleton with silhouettes in a shared representation space allows the model to capture both global shapes and fine-grained structures more effectively.

\section{CONCLUSION}
We propose Parsing Skeleton, a novel gait representation that encodes fine-grained body dynamics via skeleton-guided parsing and transforms them into image form for CNN-based feature extraction. Building on this, PSGait integrates Parsing Skeletons with silhouettes to enhance robustness and individual differentiation in real-world scenarios. Experiments across multiple benchmarks show that PSGait achieves state-of-the-art performance while remaining lightweight. These results confirm the \textbf{effectiveness}, \textbf{generalization}, and \textbf{lightweight} of Parsing Skeleton, establishing PSGait as a practical solution for multimodal gait recognition. As a general human representation, Parsing Skeleton may also provide inspiration for broader human motion analysis.

\section{ACKNOWLEDGEMENTS}
This work is supported by the NSFC fund (62576190), in part by the Shenzhen Science and Technology Project under Grant (KJZD20240903103210014, JCYJ20220818101001004)

\begingroup
\footnotesize   
\bibliographystyle{IEEEbib}
\bibliography{refs}
\endgroup

\end{document}